\documentclass[letterpaper]{article} 
\usepackage{aaai24}  
\usepackage{times}  
\usepackage{helvet}  
\usepackage{courier}  
\usepackage[hyphens]{url}  
\usepackage{graphicx} 
\usepackage{natbib}  
\usepackage{caption} 
\usepackage{algorithm}
\usepackage{algorithmic}
\usepackage{amsmath,amssymb,amsfonts}
\usepackage{multirow}
\usepackage{newfloat}
\usepackage{listings}
\DeclareCaptionStyle{ruled}{labelfont=normalfont,labelsep=colon,strut=off} 
\floatstyle{ruled}
\newfloat{listing}{tb}{lst}{}
\floatname{listing}{Listing}
\title{Learning Persistent Community Structures in Dynamic Networks \\via Topological Data Analysis}
\author {
    Dexu Kong,
    Anping Zhang,
    Yang Li\textsuperscript{\footnote{Corresponding author.}}
}
\affiliations {
    Shenzhen Key Laboratory of Ubiquitous Data Enabling, Shenzhen International Graduate School, Tsinghua University\\
    kdx21@mails.tsinghua.edu.cn,  zap21@mails.tsinghua.edu.cn, yangli@sz.tsinghua.edu.cn
}

\usepackage{bibentry}

\begin{document}

\maketitle

\begin{abstract}
Dynamic community detection methods often lack effective mechanisms to ensure  temporal consistency, hindering the analysis of network evolution. In this paper, we propose a novel deep graph clustering framework with temporal consistency regularization  on   inter-community structures, inspired by 
the concept of minimal network topological changes within short intervals.
Specifically, to address the representation collapse problem, we first introduce \textit{MFC}, a matrix factorization-based deep graph clustering algorithm that preserves node embedding. Based on static clustering results, we construct probabilistic community networks and compute their persistence homology, a robust topological measure, to assess structural similarity between them. Moreover, a novel neural network regularization \textit{TopoReg} is introduced to ensure the preservation of topological similarity between inter-community structures over time intervals. Our approach enhances temporal consistency and clustering accuracy on real-world datasets with both fixed and varying numbers of communities. It is also a pioneer application of TDA in temporally persistent community detection, offering an insightful contribution to field of network analysis. Code and data are available at the public git repository: https://github.com/kundtx/MFC\_TopoReg
\end{abstract}

\section{Introduction}

Community detection on dynamic networks is crucial for graph analysis. The formation of social ties, economic transactions, the unfolding of human mobility and communication, such real-world events all lie in the identification of meaningful substructures and their evolution hidden in the temporal complex system. Static community detection algorithms are well-researched and developed, such as the Louvain method \cite{Blondel_2008}, submodularity \cite{liu2013entropy}, and spectral clustering \cite{spectralclustering}.
As graph neural networks have shown super capabilities in fields such as node classification and link prediction, deep graph clustering methods have come to the fore \cite{zhou2022comprehensive} and been gradually adopted in static community detection \cite{su2022comprehensive}. However, dynamic community detection methods are still slow to develop due to the lack of a clear definition of communities in dynamic networks. Some dynamic community detection algorithms apply improved static algorithms on each snapshot of the network \cite{javed2018community}. Nevertheless, these methods focus on snapshot optimal solutions and their results lack consistency in time.
\begin{figure}[] 
\centering 
\includegraphics[width=0.45\textwidth]{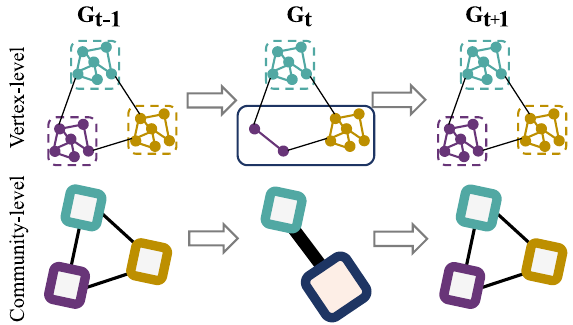}
\caption{Inconsistent inter-community structure in dynamic community detection. The top row shows three snapshots of a dynamic graph constructed at the vertex level, undergoing a transient perturbation; Different node colors represent their true community labels.  The inconsistent communities are outlined by rectangles.   The second row shows   the corresponding community-level networks, which exhibit a falsely detected merge.} 
\label{Fig.motivation} 
\end{figure}

In many dynamic community detection scenarios, it is reasonable to assume that changes in the relationship between communities occur smoothly and the inter-community structure remains relatively stable over time. Research has shown that the  structure of the network itself does not change significantly over a short period of time \cite{Corne2010}.  
Moreover, the lack of temporal consistency in dynamic community detection makes it difficult to distinguish real community evolution from network perturbation, posing challenges for subsequent matching and analysis. 
For example, Fig. \ref{Fig.motivation} illustrates an inconsistent community detection results in a simple network with three true communities, e.g. social groups. In the second time step, some perturbations, such as temporary changes in membership or interactions, would cause algorithms to incorrectly infer a merging event between two communities even though they return to being relatively independent in the next time step. A good dynamic community detection algorithm can correct the clustering results based on information from nearby snapshots. 
Similar motivations can be found in previous work such as ESPRA \cite{espra}, which introduces quantum physics to model graph perturbations. While they try to smooth the structural perturbations at the vertex level, we focus on the stability at the community level. The community networks shown in the second row of Fig. \ref{Fig.motivation} is a good model of the inter-community structure, where each node is a community and the weights of the edges are equal to the sum of the weights of the inter-community edges. Focusing on the community-level structure has a more direct impact on clustering results than simply modifying individual nodes and edges. In this paper, we investigate how to constrain the structural consistency of community networks within nearby snapshots.

We believe that the key to distinguishing inter-community structures lies in the topological characteristics of community networks. For example, in Fig. \ref{Fig.motivation}, the difference between a triangle and a line illustrates the difference in the structure of community networks. 
Since 2009, there have been increasing research efforts on Topological Data Analysis (TDA), which integrates algebraic geometry, computational geometry and data mining \cite{carlsson2009topology}. TDA characterizes intrinsic, topological changes in graph data through persistence homology, which quantify the topological features in the data across continuous scales. Topological graph analysis is a special class of TDA for graph data. Unlike heuristically designed topological features, such as the RA index \cite{Zhou2009}, persistence homology is much more scale-independent and robust to perturbations, making it a better choice for quantifying the structural similarity in community networks.

In this work, we propose  a novel dynamic community detection framework, which jointly performs graph clustering at the vertex-level  and temporal consistency regularization at the community-level.  Specifically, we  solve  two  main challenges. 
First, the widely used self-supervised clustering module \cite{xie2016unsupervisedDEC} would collapse local structure of embedding distribution \cite{IDEC2017}.
To preserve the structure of the embedding space, we propose a novel deep graph clustering algorithm called MFC, which inspired by non-negative matrix factorization.
The second challenge lies in how to incorporate the TDA-based consistency constraints on the community-level structures with vertex-level graph clustering methods. In this work, we design a differentiable operator that associates topological features with the cluster assignment distribution in deep clustering algorithms. Thus, the gradient of our Topological Regularization (TopoReg) can be back-propagated to the clustering module to penalize the topological differences in the community structure in neighboring snapshots. 
  
The main contributions of our work are concluded as follows:
\begin{itemize}
    \item We introduce topological graph analysis into dynamic community detection to learn consistent inter-community structures end-to-end.
    \item We propose a novel deep clustering algorithm that implements matrix factorization with relaxed sparsity constraints via neural networks.
    \item We empirically demonstrate the superiority of our proposed clustering algorithm and the importance of community structure preservation in dynamic community detection with both fixed and varying numbers of communities.
\end{itemize}

\section{Related Works}

\subsection{Deep Graph Clustering}

Deep graph clustering, clustering nodes in a graph into communities, is an emerging field in machine learning and social networks. 
We divide existing deep graph clustering methods into two classes: Static graph clustering or Dynamic graph clustering.

\noindent\textbf{Static graph clustering.} Most frameworks perform clustering on lower-dimensional embedding of graphs, based on popular architectures like GANs \cite{creswell2018generative}, and Graph Auto-Encoders \cite{kipf2016variational}. A naive approach would directly perform traditional community detection methods on node embedding. EGAE \cite{zhang2022embedding}, a work very similar to our approach, is a typical one. It finds an ideal space for the clustering, but still uses $k$-means. Instead, another self-optimized deep clustering framework jointly optimizes the learned embedding and perform clustering, such as DAEGC \cite{wang2019attributedDAEGC}. Their core clustering module comes from DEC \cite{xie2016unsupervisedDEC}. Recent models improve deep graph clustering by better learning of node features, i.e. SCDN \cite{bo2020structural}, AGCN \cite{peng2021attention} and DCRN \cite{dcrn}, but the core clustering module does not change.

\noindent\textbf{Dynamic graph clustering.} Although there have been successful studies on evolving graphs, they mostly focus on node classification \cite{pareja2020evolvegcn} or temporal networks clustering, which typically yields a single clustering result for networks with changing edge weights \cite{liu2023deep}. In contrast, this paper focuses on tracking community changes over discrete snapshots, which is relatively underexplored in the field. Traditional dynamic community detection algorithms, such as RTSC \cite{you2021robust}, 
ESPRA \cite{espra}, DECS \cite{liu2020detecting}, often solve a multi-objective optimization problem. 
CGC \cite{park2022cgc} improves the graph clustering algorithm based on contrastive learning and extends it to dynamic graphs, but its experiments are performed on a dataset with binary labels only. Dynamic graph embedding algorithms combine recurrent neural network with graph autoencoders, such as DynAE, DynRNN, and DynAERNN \cite{goyal2020dyngraph2vec}. Though without a dedicated clustering module, some clustering functionality is available. 

\subsection{Topological Graph Analysis}
TDA can be extended to graphs by representing them as simplicial complexes, which encode their topology and structural properties. Numerous graph filtration methods have been proposed methods to compute persistent homology of graphs, such as Vietoris-Rips filtration \cite{dey2022computational}, weighted simplex filtration \cite{huang2016persistent}, and vertex-based clique filtration \cite{rieck2017clique}. The graph topological features extracted from these graph filtration methods are widely used in biological and social graph data, among others. A series of studies on brain networks using TDA were presented by Songdechakraiwut and Chung \cite{songdechakraiwut2023topological}, such as learning MRI signals by graph filtration to understand complex relations in brain networks. Periodic phenomena in temporal traffic networks were studied using WRCF by Lozeve \cite{lozeve2018topological}, while Hajij \cite{Hajij2018} uses Rips filtration to visualize structural changes in dynamic networks. In addition, recent works \cite{yan2021link} have shown that adding topological graph analysis into GNNs can effectively improve their learning ability and performance.

\section{Notations and Problem Formulation}

Given a graph $G=(V, E)$, $V$ is a set of vertices and $E$ is a set of tuples $\left(u, v\right)$ with $u, v \in V$. A Dynamic Graph $\mathcal{G}_\tau$ is defined by an ordered set $\{ G^{(1)}, G^{(2)}, \ldots, G^{(t)}\}$. Dynamic community detection is to find the best cluster assignment $Y^{(t)}$ for each snapshot $G^{(t)}$ at time step $t$.
The criteria we define for good dynamic community detection are twofold: on the one hand, we need to achieve cohesive clustering results at each snapshot, and on the other hand, we expect the structure of the detected communities to maintain a certain degree of topological stability and continuity during dynamic changes of the input network. Our methods provide a trade-off between snapshot coherence and temporal consistency.

\section{Methodology}

In this section, we present our topology preserving dynamic community detection framework in detail. First, we will  start by introducing a novel static deep graph clustering algorithm, followed by the topological consistency regularization for communities derived from clustering result within neighboring time windows. Note that we omit the time dimension in the Matrix Factorization Clustering section to make the notations more readable.

\subsection{Matrix Factorization Clustering}
Node embeddings learned by common deep clustering method tend to collapse to cluster centroids, which is desirable for node label inference but difficult to train in the presence of regularization. For example, DEC works by taking the proximity between the node embeddings and each cluster centers in the embedding space as the cluster assignment distribution. By constraining the sparsity of the distribution, the samples in each cluster are clustered toward the center. However, there is no guarantee that the samples near the margin will be pulled into the correct cluster, and it will destroy the structure of embedding space to some extent. Therefore, we propose a novel end-to-end deep graph clustering algorithm. It is called Matrix Factorization Clustering (MFC), a novel end-to-end deep graph clustering algorithm that learns relaxed matrix factorization on node embeddings using a Graph Auto-encoder (GAE). In contrast to DEC, MFC is a dimension reduction technique that avoids lossy compression, thereby preserving the structure of the embedding space.
GAE incorporates the traditional auto-encoder and different kinds of GNNs, such as GCNs \cite{zhang2019graph} and GAT \cite{salehi2020graph}. Like a common auto-encoder, GAE consists of an encoder and a decoder. The encoder part attempts to learn a latent representation $Z=[\boldsymbol{z}_1,\boldsymbol{z}_2,...,\boldsymbol{z}_n]^\mathrm{T}$ of graph input with $n$ nodes via GNN layers. While the decoder intends to reconstruct the adjacency matrix $A$ from embedding $Z$, it is usually designed as $\sigma(ZZ^\mathrm{T})$ where $\sigma(\cdot)$ denotes the sigmoid function. The reconstruction loss $\mathcal{L}_{gae}$ is the binary cross entropy loss between $A$ and $\sigma(ZZ^\mathrm{T})$. 
In general, the clustering results can be obtained by directly performing $k$-means or other heuristic algorithms on embedding $Z$. However, the clustering results obtained in this way are not differentiable, so it is difficult to further optimize the topology of the detected community structure.   

\begin{figure}[] 
\centering 
\includegraphics[width=0.48\textwidth]{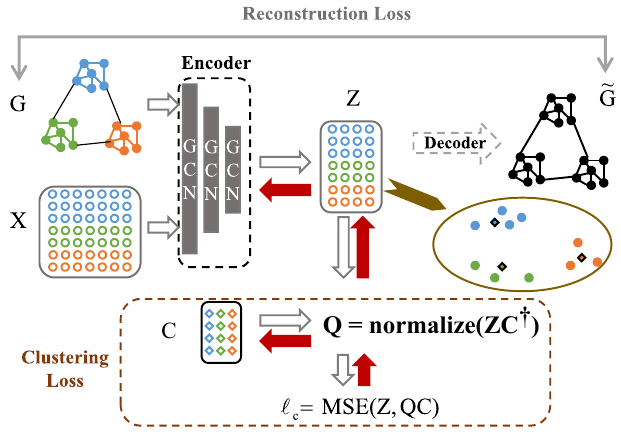} 
\caption{Framework of matrix factorization clustering. It consists of a graph auto-encoder and a clustering module.} 
\end{figure}

Matrix factorization has been proven to be essentially equivalent to $k$-means, spectral clustering, and many other clustering algorithms \cite{math11122674}. In our method, with relaxed sparsity constraints, Neural networks can be used to learn two low-rank matrices decomposed from the embedding matrix. Specifically, the optimization problem solved by $k$-means can be expressed as:

\begin{equation}
\begin{aligned}
&\min _{C, Q}\left\|Z-Q C\right\|_F^2, \\
&\text { s.t. } q_{i j} \in\{0,1\},Q\mathbf{1}_k=\mathbf{1}_n.
\end{aligned}
\end{equation}
where $Q=(q_{ij})$ and $C=[\boldsymbol{\mu}_1,\boldsymbol{\mu}_2,...,\boldsymbol{\mu}_k]^\mathrm{T}$. $\boldsymbol{\mu}_j$ denotes center of $j$ cluster and $q_{ij}$ is the indicator. Specifically speaking, $q_{ij} = 1$ if the $i$-th point is assigned to the $j$-th cluster. Otherwise, $q_{ij} = 0$. The above problem is hard to solve directly due to the discrete constraint on $Q$. We first derive the closed-form solution of $Q$ in the unconstrained situation when $C$ is fixed. The objective function can be derived as: 
\begin{equation}
    \begin{aligned}
    \mathcal{J}_{km} & =\left\|Z-Q C\right\|_F^2 \\
    & =\operatorname{tr}\left(Z^\mathrm{T} Z\right)-2 \operatorname{tr}\left(Z^\mathrm{T} Q C\right)+\operatorname{tr}\left(C^\mathrm{T} Q^\mathrm{T} Q C\right) .
    \end{aligned}
\end{equation}
Take the derivative of $\mathcal{J}_{km}$ and set it to 0
\begin{equation}
    \begin{aligned}
        &\nabla_Q \mathcal{J}_{km} = 2(QCC^\mathrm{T} - ZC^\mathrm{T})=0 \\
        &Q = ZC^\mathrm{T}(CC^\mathrm{T})^{-1}=ZC^{\dagger}
    \end{aligned}
\end{equation}
Inspired by Pseudo Inverse Learning \cite{GUO2004101} and ADMM (Alternating Direction Method of Multipliers) \cite{admmboyd}, we consider $C$ as a set of weights in the neural network, thus combining the alternating optimization of $C$ and $Q$ with the training process. Specifically, $Q$ is updated by $Q=g(ZC^{\dagger})$ in forward propagation, the encoder of GAE and $C$ are updated by gradient descent in backward propagation. Here $g$ is a function that project $Q$ to the feasible region. Specifically, we relax the discrete constraints on $Q$ to a soft assignment problem, which means $ q_{i j} \in(0,1), \sum_{j=1}^k q_{ij}=1$.  Normalized by Softmax, Min-Max Normalization or other algorithms, any continuous matrix can satisfy the condition. In this paper we choose Min-Max Normalization to normalize each row $q$ of $Q$ as the relaxed constraints:
\begin{equation}
g(q)=\frac{q-\min (q)}{\max (q)-\min (q)}
\end{equation}

The MSE (Mean Square Error) between $Z$ and $g(Q)C$ is calculated as $\mathcal{L}_{c}$ to jointly train the neural network with $\mathcal{L}_{gae}$ using back propagation until convergence. The detailed
learning procedure of MFC is shown in Algorithm \ref{alg:algorithm}. When the best clustering centers $C^*$ is learned, the index of row maximum in $Q^*$ can be taken as the final clustering result of each node: $y_i = \arg \max_u q_{iu}$. One limitation of MFC is that since the principle of the method is matrix decomposition, the algorithm fails when the dimension of the node embeddings is less than the number of clusters. Because when the dimension of a matrix is greater than its rank, its pseudo-inverse does not exist. 

\begin{algorithm}[tb]
    \caption{Optimizing clustering module in MFC}
    \label{alg:algorithm}
    \textbf{Input}: Node embedding matrix $Z$ of graph $G$ learned by GAE\\
    \textbf{Parameter}: Trade-off parameter $\alpha$\\
    \begin{algorithmic}[1] 
        \STATE Initialize weights $C$ .
        \REPEAT
        	\STATE Calculate $Q=ZC^{\dagger}$.
            \STATE Normalize each row of $Q$ with function $g$.
            \STATE Calculate the differentiable clustering loss $\mathcal{L}_{c}=MSE(Z, g(Q)C)$ .
            \STATE Calculate the total loss and its gradients: $\mathcal{L}_{gae} + \alpha \times \mathcal{L}_{c}$.
            \STATE Update GAE weights and $C$ by gradient descent.
        
        \UNTIL convergence or exceeding maximum iterations

        \STATE \textbf{Ensure} Assignment Matrix $Q$, clustering centers $C$  and GAE encoder parameters $\{W_i\}_{i=1}^L$
    \end{algorithmic}
\end{algorithm}

\subsection{Topological Regularization of Dynamic Clustering Consistency}

\begin{figure}[] 
\centering 
\includegraphics[width=0.48\textwidth]{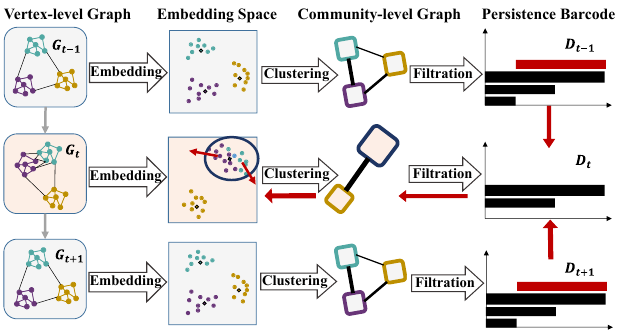} 
\caption{Illustration of topological regularization. 
A sequential process across three time steps is shown. Rows trace graph evolution with columns showing snapshots, embeddings, community graphs, and persistence barcodes. Different colors are used to differentiate the real category labels. Red arrows highlight the temporal consistency loss on persistence barcodes and the backpropagation path for the $t$th snapshot $G_t$. 
} 
\label{Fig.topoloss} 
\end{figure}

We propose an end-to-end regularization TopoReg to ensure the topological consistency of the community networks, which makes the clustering results more accurate and stable. 
It is a sliding window style loss function in order to reduce the topological distance between neighboring community networks. Furthermore, to feed the gradient back into the deep graph clustering module, we devise an elaborated community network construction method to make the community topology a differentiable function of cluster membership. The complete process is shown in Fig.\ref{Fig.topoloss}.

\noindent\textbf{Construction of Community Topology.}
Given a graph $G$, the cluster assignment distribution $Q$ is computed in most deep graph clustering method such as DAEGC and the MFC algorithm we proposed. We can always assign a pseudo label $s_i$ on each node $i$ based the index of maximum row value of $Q$. This pseudo division of $G$ could form a new graph containing community structure called community network, whose nodes are communities. The weight of an edge connecting two communities A and B is determined by summing up the weights of all the edges that have one end belongs to A and the other belongs to B. The following equations demonstrate the derivation of community network based on $Q$ and the weighted adjacency matrix $W$ of $G$. 

The pseudo label of node $i$ is $s_i = \arg\max_k(Q_{ik}), \forall i \in \{1, 2, \ldots,n\}$.  Assuming that graph $G$ has $K$ ground truth clusters, we have $s_i\in \{1,2,\dots,K\}$.  Organize the pseudo-labels of all nodes into a vector $S=[s_1, s_2,\cdots, s_n]$, we define the indicator function  $ \mathbf{1}\left(S=k\right)=[\mathbf{1}\left(s_1=k\right),\mathbf{1}\left(s_2=k\right), \ldots, \mathbf{1}\left(s_n=k\right)]^\mathrm{T}=[0,0, \ldots 1, \ldots 0]^\mathrm{T}$, where $\mathbf{1}\left(s_i=k\right)=1. \iff  s_i=k$. If we take the $k$th column of $Q$ called $Q_{k}=[q_{1 k}, q_{2 k}, \cdots, q_{n k}]^\mathrm{T}$, 
 a filtered distribution matrix $\hat{Q}_{k}=Q_{k} \odot \mathbf{1}\left(S=k\right)$, which means the removal of entries in the $k$th column that are not row maxima. We define edge weight between community $1$ and community $2$ as:
\begin{equation}
\begin{aligned}
    M_{1 2}
    & =\Sigma_j \Sigma_i \mathbf{1}\left(s_i=1\right) \cdot q_{i 1} \cdot w_{i j} \cdot \mathbf{1}\left(s_j=2\right) \cdot q_{j 2} \\
    & =\Sigma_j \Sigma_i \hat{q}_{i 1} w_{i j} \hat{q}_{j 2} \\
    & =\hat{Q}^\mathrm{T}_{1} W \hat{Q_{2}}\\
\end{aligned}
\end{equation}

For intuition, assume that $Q$ is a discrete matrix in which each row includes only one $1$ and the rest are $0$s. In this scenario, $M_{1 2}$ equals the number of edges between the two communities. If we organize $\hat{Q}_{k}$ in to a matrix $\hat{Q}= [\hat{Q}_1, \hat{Q}_2, \dots, \hat{Q}_K]$, then the adjacency matrix $M \in \mathbb{R}^{K \times K}$ of the community network can be written as: $\hat{M}=\hat{Q}^\mathrm{T} W \hat{Q}$. To adapt to networks of different sizes, we need to normalize $\hat{M}$ into $M$:

\begin{equation}M =\frac{\hat{Q}^\mathrm{T} W \hat{Q}}{\sum_i\sum_j W_{ij}},\end{equation}
where the denominator is the sum of the weights of all edges. Based on the construction method mentioned earlier, each edge in the community graph equals the sum of the products of the assignment probabilities that the endpoints of the edge connecting corresponding communities. Given any graph filtration $f$, the persistence diagram of community network is $\mathbf{dgm}( M)$, which quantifies topological characteristics of community networks.  Since the Betti numbers and graph weights can establish a one-to-one correspondence,  the gradient of the loss function based on the Wasserstein distance can be backpropagated to the parameters of the graph encoder, which changes the node embedding $Z$. Thus the clustering results are optimized to ensure a persistent community topology.

\begin{figure}[] 
\centering 
\includegraphics[width=0.45\textwidth]{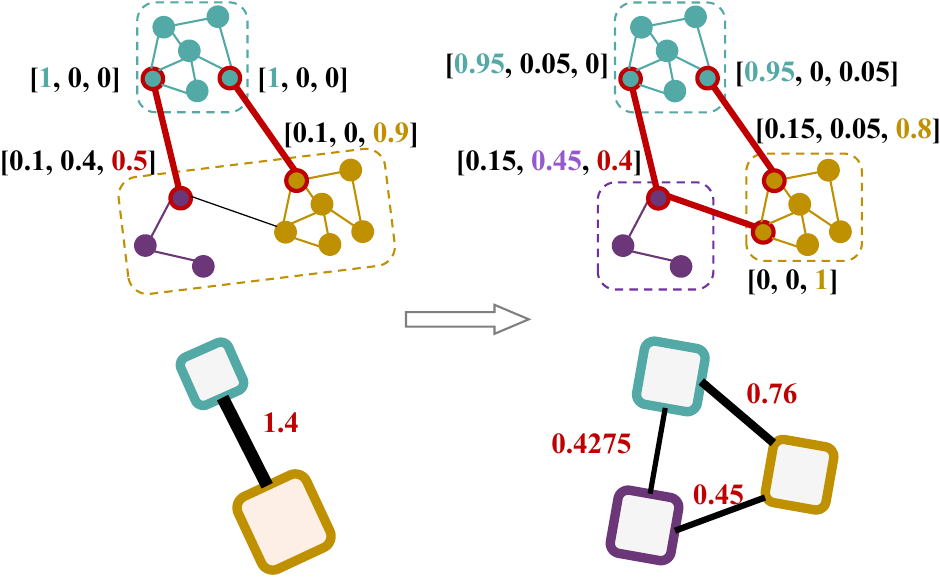} 
\caption{A demo showing how the community graph is computed. The first row shows the clustering results, and the second row shows the community graphs derived from them. From left to right, the assignment distribution of the nodes marked in red changes as the gradient of the edge weight of the community graph decreases, and the corresponding topology changes.} 
\label{Fig.comgraph} 
\end{figure}

\noindent\textbf{Topological Loss Definition.}
In our method, we perform Weight rank clique filtration (WRCF) \cite{petri2013topological} on the community network to calculate the \textit{$0$}-th and \textit{$1$}-th Persistence Diagram (PD) of the community topology. They record the Betti numbers $\beta_0$ and $\beta_1$ , respectively, reflecting the connected components and the two-dimensional voids. WRCF sequentially adds edges with higher weights to form simplicial complexes. This technique identifies maximal cliques based on the subgraph at each filtration level for topological analysis and facilitates the application of persistent homology to study structural changes over time. Given a dynamic graph  $\mathcal{G}_\tau=\{G^{(t)}\}_{t=0}^T$, we apply deep graph clustering on each snapshot and compute their community networks $\{M^{(t)}\}_{t=0}^T$ based on clustering assignment distribution matrices. Then WRCF are applied to them to get a series of PDs $\{\mathbf{dgm}(M^{(t)})\}_{t=0}^T$. By calculating the Wasserstein distance \cite{carriere2017sliced} between the PD at the current snapshot and the PDs before and after, we can construct a constraint on the consistency of the clustering results, formulated as:

\begin{equation}
\label{eq:loss topo}
    \begin{aligned}
        \mathcal{L}_{topo}  = 
        &\sum_{t=1}^{T-1} \sum_{k\in\{1,2\}}  \left( \mathrm{W}_{p, q}\left(\mathbf{dgm}_k(M^{(t)}),\mathbf{dgm}_k(M^{(t-1)})\right)\right.\\
        &\left.+\mathrm{W}_{p, q}\left(\mathbf{dgm}_k(M^{(t)}), \mathbf{dgm}_k(M^{(t+1)})\right) \right).\\
    \end{aligned}
\end{equation}

One technicality is that the two diagrams may have different cardinalities. In this case,  some extra points will be mapped to the diagonal line in Wasserstein distance. In practice, we choose $p=1$ and $q=\infty$, which corresponds to 
Earth Mover distance and the Infinity-norm, respectively. 
The complete training process of the model is summarized in Algorithm \ref{alg:algorithm2}.
\begin{algorithm}
\caption{Complete Training Process}
\label{alg:algorithm2}

\begin{algorithmic}[1]
\REQUIRE A dynamic graph $\mathcal{G}_\tau=\{G^{(t)}\}_{t=0}^T$ and trade-off hyper-parameter $\alpha$
\FOR{$t = 1$ to $T$}
    \STATE Optimize the GAE on the snapshot $G^{(t)}$ to obtain node embedding $Z^{(t)}$ by minimizing the reconstruction loss $\mathcal{L}_{gae}$.
    \STATE Perform MFC on the embedding $Z^{(t)}$ to learn clustering assignment $Q^{(t)}$ by optimizing the composite loss $\mathcal{L}_{gae} + \alpha \mathcal{L}_{c}$.
    \STATE Compute community network $M^{(t)}$ based on $Q^{(t)}$
    \STATE Compute persistence diagrams $\mathbf{dgm}(M^{(t)})$
\ENDFOR
\STATE Integrate topological insights by calculating the Topological Loss $\mathcal{L}_{topo}$, and apply backpropagation to refine the model.
\end{algorithmic} 
\end{algorithm}

\begin{figure}[] 
\centering 
\includegraphics[width=0.45\textwidth]{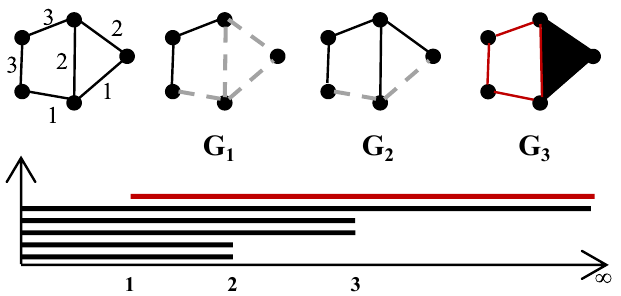} 
\caption{Illustration of Weight Rank Clique Filtration (WRCF) applied to a toy graph. The first row shows a weighted graph, followed by the simple complex under three levels of filtration, and the second row shows the persistence barcode corresponding to the above filtration. The black lines represent the 0th persistent Betti number, while the red line represents the 1st persistent Betti number.} 
\label{Fig.wrcf} 
\end{figure}

\noindent\textbf{Topological Optimization.} To introduce topological loss into the deep learning framework, we need to calculate its gradient. Computation of persistence homology is typically based on non-continuous matrix reduction algorithms. Since the output is in the form of a multiset, calculating the gradient directly poses difficulties. Following \cite{gabrielsson2020topology}, given a graph filtration, each birth-death pair in persistence diagrams can be mapped to the edges that respectively created and destroyed the homology class.
If the ordering on simplices is strict, the map will be unique, and we can obtain the gradient by inverse mapping the birth-death values to edge weights. Note that if the ordering is not strict, which is more likely, we can still extend the total order to a strict order either deterministically or randomly. 

\section{Experiments and Results}
In this section, experiments on both synthetic data and real-world datasets are conducted to evaluate the performance and effectiveness of our algorithm. We compare the performance of our algorithm with the state-of-the-art algorithms. The experiment is repeated five times and the average results are reported to account for any variation in results.
\begin{figure}[] 
\centering 
\includegraphics[width=0.48\textwidth]{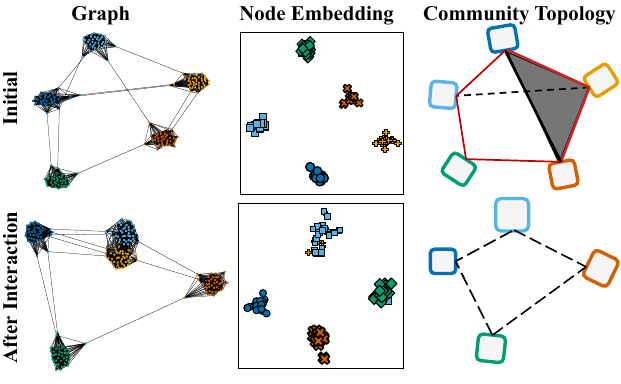} 
\caption{Visualization of the demo graphs with their embedding and community topology. These three lines show graph, embedding, and community topology respectively. Different markers and colors represent different real clusters. The triangle in $(e)$ represents a 2-simplex and the red lines highlight a $\beta_1$} 
\label{Fig.graph_emb_topo} 
\end{figure}

\subsection{Experiments on Synthetic Data}
We reproduce a similar scenario in Fig.\ref{Fig.motivation}: there are five groups of people, and two of them have additional links due to an ephemeral collaboration. We will show that TopoReg successfully ensures the temporal structure consistency of the community detection results. In Fig.\ref{Fig.graph_emb_topo}, a Gaussian random partition graph is initialized by creating 5 clusters each with a size drawn from a normal distribution $\mathcal N(20,1)$. Nodes are connected within clusters with a probability of 0.5 and between clusters with a probability of 0.001. Then, the other graph is created by randomly adding moderate amount of edges between 2 of the 5 clusters. Their node embeddings are visualized in the middle column respectively by dimension reduction to 2D via t-distributed stochastic neighbor embedding (t-SNE) \cite{van2008visualizing}. The third column shows that the collaboration leads to a sudden change in the community topology. Specifically, two 2-cliques are wiped out. The initial graph has 4 $\beta_0$ and 2 $\beta_1$ features, whereas the graph after interaction has only 3 $\beta_0$  and no $\beta_1$, since 2-simplex does not exist. 

In such a situation, our algorithm is shown to have a binding influence on the node embedding, thus changing the clustering results and maintaining a stable topology between communities. We cluster the two graphs separately and optimize the clustering result of the second graph with $\mathcal{L}_{topo}$. In Fig. S1 (Appendix), we track the changes in node embedding, topological loss, and clustering effect. We find that the two clusters of points, which were initially more concentrated, gradually separate as the loss decreases. The clustering accuracy also improves from 81\% to 98\%.

\subsection{Experiments on Real-world Datasets}
\noindent\textbf{Datasets} We collected and processed four labeled dynamic network datasets without node features, including Enron, Highschool \cite{crawford2018cluenet}, DBLP, Cora \cite{hou2020glodyne}.  Noting that each node in these existing datasets has a fixed label, we processed a new dataset DBLP$_{dyn}$ from the original data, recalculating the node's label at each snapshot. The brief information of these datasets is summarized in Table S1. 

\begin{figure}[]
\centering 
\includegraphics[width=0.48\textwidth]{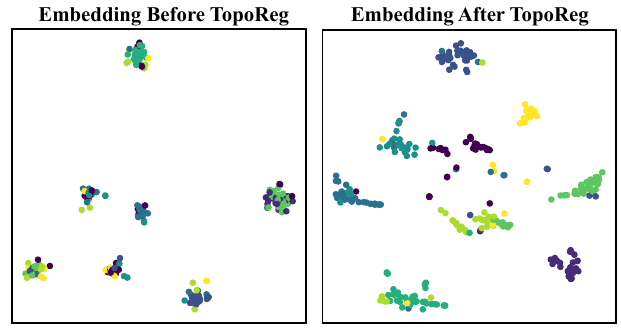} 
\caption{A comparison of node embeddings before and after applying TopoReg to GEC algorithm. The data is obtained from a single snapshot in the Highschool dataset, and the colors indicate the ground truth of community labels.}
\label{Fig.highschoolemb}
\end{figure}
\begin{table*}[h]
\begin{center}
\resizebox{\textwidth}{!}{
\begin{tabular}{cc|ccc|ccc|ccc|c}
\hline
\multirow{2}{*}{Data}       & \multirow{2}{*}{Metrics} & \multicolumn{3}{c|}{Static Baselines} & \multicolumn{3}{c|}{Temporal Baselines} & \multicolumn{3}{c|}{Ablation Study}    & Ours           \\ \cline{3-12} 
                            &                          & GEC         & DAEGC      & SDCN       & DECS           & ESPRA & DynAE          & MFC         & GEC+Topo    & DAEGC+Topo & MFC+Topo       \\ \hline
\multirow{4}{*}{Enron}      & ACC                      & 58.66       & 58.15      & 57.86      & 57.24 & \underline{59.82} & 58.49          & 58.44       & \textbf{60.32} & 58.33      & 59.31          \\
                            & NMI                      & 15.42       & 15.69      & 12.44      & 15.63    & 13.85 & 7.10            & 18.9        & \underline{18.14}       & 17.53      & \textbf{19.14} \\
                            & ARI                      & 0.47        & -0.81      & -1.10       & -1.90          & -0.30 & 1.47           & \underline{2.17}  & 1.00           & -0.36      & \textbf{2.73}  \\
                            & Modularity               & 30.42       & 30.08      & -1.84         & 45.40          & -2.10 & -1.47          & \underline{45.54} & 39.34       & 39.18      & \textbf{46.36} \\ \hline
\multirow{4}{*}{Highschool} & ACC                      & 49.21       & 49.33      & 24.02      & 65.77 & 26.44 & 18.82          & \underline{68.91}       & 63.51       & 62.66      & \textbf{70.67}    \\
                            & NMI                      & 28.11       & 42.58      & 9.71       & 62.36    & 12.31 & 5.64           & 63.14       & 59.41       & 56.22      & \textbf{65.78} \\
                            & ARI                      & 13.57       & 26.43      & 0.46       & 36.18 & 0.12  & 0.11           & \underline{48.77}       & 44.44       & 38.93      & \textbf{50.87}    \\
                            & Modularity               & 49.82       & 56.35      & -0.25      & 72.80 & -0.95 & -0.11          & \underline{76.99}       & 73.62       & 68.37      & \textbf{77.87}    \\ \hline
\multirow{4}{*}{DBLP}       & ACC                      & 56.38       & 56.23      & 56.22      & OOM            & OOM   & \textbf{68.31} & 56.83       & 57.62       & 56.54      & \underline{57.98}    \\
                            & NMI                      & 1.75        & 2.41       & 1.57       & OOM            & OOM   & 0.28           & \underline{6.32}  & 6.31        & 3.92       & \textbf{7.97}  \\
                            & ARI                      & 0.25        & 0.35       & 0.37       & OOM            & OOM   & 0.05           & 0.98        & \underline{1.37}  & 0.62       & \textbf{1.41}  \\
                            & Modularity               & 56.07       & 59.28      & 4.54       & OOM            & OOM   & 0.07           & \underline{85.39} & 76.71       & 71.61      & \textbf{86.65} \\ \hline
\multirow{4}{*}{Cora}       & ACC                      & 35.18       & 37.56      & 34.17      & OOM            & OOM   & 37.85          & \underline{50.66} & 43.18       & 41.6       & \textbf{52.53} \\
                            & NMI                      & 3.21        & 6.79       & 1.45       & OOM            & OOM   & 0.24           & \underline{24.27} & 16.10        & 11.66      & \textbf{27.52} \\
                            & ARI                      & 1.18        & 3.05       & 0          & OOM            & OOM   & 0.18           & \underline{13.49} & 8.04        & 5.50        & \textbf{14.62} \\
                            & Modularity               & 47.70       & 49.94      & 3.59       & OOM            & OOM   & 1.32           & \underline{74.09} & 63.05       & 55.53      & \textbf{75.61} \\ \hline
\end{tabular}
}

\end{center}
\caption{Experimental results on four datasets with known cluster numbers. $K$-means clustering is performed on graph embedding to obtain the final community membership. OOM means out-of memory.}
\label{table:exp}
\end{table*}

\noindent\textbf{Baselines}

\begin{itemize}
    \item \textbf{Static Baselines.} We first compare with state-of-the-art deep graph clustering method focusing on static community detection. DAEGC \cite{wang2019attributedDAEGC} first introduces the clustering module in DEC \cite{xie2016unsupervisedDEC} into the graph clustering problem. In order to simplify the problem and make the comparison fair, we also compare with the version that replaces GAT back to basic GCN in encoder, leaving other parts unchanged, which is called Graph Embedding Clustering (GEC). SDCN \cite{bo2020structural} improves deep clustering by integrating the structual information into representation learning module, but the core clustering module remain the same. 
    \item \textbf{Temporal Baselines.} ESPRA \cite{espra}, DECS \cite{liu2020detecting} are two local smoothing dynamic community detection methods. Since they solve a multi-objective optimization problem with graph data in the form of a 3D matrix, they are too time- and memory-consuming and cannot be run on DBLP and Cora datasets. DynAE, DynRNN, DynAERNN \cite{goyal2020dyngraph2vec} are a family of dynamic graph embedding algorithms that use recurrent neural networks to model temporal information in dynamic networks. DynRNN, DynAERNN expand network parameters based on DynAE. Limited by GPU memory, we only report DynAE results. Note that DECS is a label propagation algorithm and cannot fix the number of clusters, so we take the top $k-1$ clusters in the clustering result and merge the remaining nodes into one cluster. 
\end{itemize}

\noindent\textbf{Metrics}  We use Accuracy (ACC), Normalized Mutual Information (NMI), Adjusted Rand Index (ARI) to reflect how well the clustering results match the ground truth labels, and Modularity to reflect the clustering quality. All metrics are calculated at each snapshot and the average values are reported in Table \ref{table:exp}.

\noindent\textbf{Experimental Settings} The experiments in this paper follow the following settings: node embedding dimension is 30, the learning rate is 0.001. Each backbone method uses one GNN layer (GCN, GAT). The other baseline models are set as default by the authors. The training is divided into two stages, $\alpha=10$, $\beta=0$ when training the backbone model and $\alpha=1$, $\beta=10$ when training the TopoReg, 500 epochs of each to ensure convergence. All weights in neural networks are initialized with Glorot initialization \cite{glorotinit}.

Note that the nodes and edges in our dataset may appear and disappear in each snapshot, so the number of nodes per snapshot is not consistent. For temporal baselines that require a fixed-size dynamic adjacency matrix input, we add the points that do not appear in the current snapshot as isolated nodes. And to be fair, we remove these isolated nodes when calculating the metrics.

\begin{figure}[]
\centering 
\includegraphics[width=0.48\textwidth]{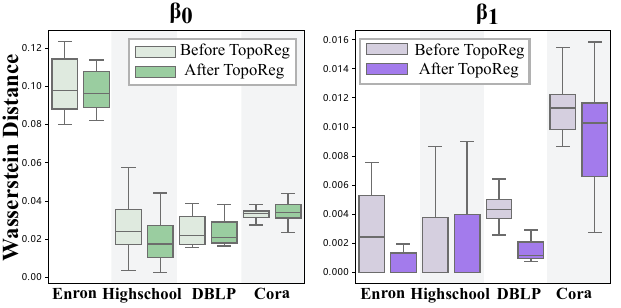} 
\caption{Comparisons of the Wasserstein distances between the community topology under the ground truth labels and the deep clustering labels before and after applying TopoReg.} 
\label{Fig.topo_shift} 
\end{figure}

\noindent\textbf{Results with Fixed Community Number} Table \ref{table:exp} shows the experimental results on four real-world datasets with constant node labels. In such a situation, we assume that the community number $k$ is known and fixed, so $k$-means clustering is performed on each learned embedding. Topologically optimized MFC consistently achieves the best (bold) or second-best (underlined) accuracy, which demonstrates the superiority of our methodology. In addition, the TopoReg has an average improvement of 11.54\%, 5.90\%, and 1.38\% on the three backbone models GEC, DAEGC, and MFC, respectively. Fig. \ref{Fig.highschoolemb} is a t-SNE visualization of node embedding on one snapshot in Highschool dataset before and after applying TopoReg to DEC. It is shown that the embedding gets scattered because TopoReg optimizes the problem of representation collapse by smoothing inter-community structure.

Meanwhile, to prove that we get a more stable community structure after using TopoReg, we computed the Wasserstein distance between the persistence diagrams of   deep clustering results and that of the ground truth communities. The distribution of the distances are shown in Fig. \ref{Fig.topo_shift}. It can be seen that the topological consistency improvement of different datasets are reflected in different ways, including a decrease in median or variance. The enhancement tends to focus on $\beta_0$ or $\beta_1$ based on network properties. Overall, the inter-community structure after TopoReg is much closer to the one in the ground truth. The consistency of the ground truth labels suggests that we have found a more stable dynamic community detection results.


\noindent\textbf{Results with Varying Community Number} Table \ref{table:Q} shows the results on DBLP$_{dyn}$ dataset. The true number of communities is not known in most cases. To solve that, a heuristic algorithm is often used to select the correct number of clusters, such as the elbow method \cite{liu2020determine}. When we set the clustering dimension $K$ to a relatively large value in the model and get the clustering result based on $\arg \max Q$. The number of clusters obtained will follow the clustering structure, and finally we get a detection result where the number of communities varies dynamically. For DEC-based backbone models, they are not as well adapted to TopoReg as our MFC in this situation. 

\begin{table}[]
\resizebox{0.48\textwidth}{!}{
\begin{tabular}{cc|ccc}
\hline
Data                     & Metrics     & GEC$_{Q}$+Topo & DAEGC$_{Q}$+Topo & MFC$_{Q}$+Topo \\ \hline
\multirow{4}{*}{DBLP$_{dyn}$} & ACC        & 39.49    & 39.06      & \textbf{42.82} \\
                         & NMI        & 2.91     & 2.13       & \textbf{9.32}  \\
                         & ARI        & 0.75     & -3.61      & \textbf{2.68}  \\
                         & Modularity & 61.27    & 34.83      & \textbf{84.71} \\ \hline
\end{tabular}
}
\caption{Experimental results on dataset with unknown cluster number. The label of the corresponding node is assigned as the index of the row maximum in cluster assignment distribution matrix.}

\label{table:Q}
\end{table}

\section{Conclusion}
This work proposes an end-to-end  framework for dynamic community detection. It uses a neural network module MFC to implement matrix factorization for clustering, which outperforms the widely used self-supervised clustering method in the absence of node features. Regularization module TopoReg optimizes the cluster assignment distribution in deep graph clustering based on the topology of nearby snapshots. We demonstrate through synthetic and real dataset experiments that TopoReg can improve dynamic graph clustering results and preserve persistent community structure in terms of its  topological features. It has good theoretical interpretability and can be easily extended to other depth graph clustering architectures. The two modules provide a trade-off between node clustering and topological stability of the community. Compared to other DEC-based backbone models such as DAEGC, TopoReg combines better with MFC in the case of unknown number of communities. At this point, we can obtain dynamically changing number of clusters based on the cluster assignment distribution learned. Using distributed computing and scalable graph representations like FastGAE \cite{fastgae}, our method could be efficiently extended to large-scale graphs, since topological regularization is not directly affected by the size of graph, but only by the number of clusters.

\section{Acknowledgments}
This study is supported in part by the Tsinghua SIGS Scientific Research Start-up Fund (Grant QD2021012C), Natural Science Foundation of China (Grant 62001266) and Shenzhen Key Laboratory of Ubiquitous Data Enabling (No.ZDSYS20220527171406015)

\bibliography{aaai24}

\end{document}


\maketitle
\appendix
\newcommand{\beginsupplement}{%
        \setcounter{table}{0}
        \renewcommand{\thetable}{S\arabic{table}}%
        \setcounter{figure}{0}
        \renewcommand{\thefigure}{S\arabic{figure}}%
     }
\renewcommand{\thealgorithm}{S\arabic{algorithm}}
\renewcommand{\figurename}{Fig.}
\beginsupplement

\section{Dataset}
We collected four labeled dynamic network datasets without node features, including Enron, Highschool \cite{crawford2018cluenet}, DBLP, Cora \cite{hou2020glodyne}. Here we introduce the detailed information of these datasets.
\begin{itemize}
    \item \textbf{Enron} dataset records the email communications of 182 employees between 2000 and 2002, with each employee having one of 7 labels representing their role in the company. The initial snapshot has 76 nodes and 161 edges, and the final snapshot has 18 nodes and 22 edges. 
    \item \textbf{Highschool} dataset corresponds to contacts between 327 students in a high school in Marseilles, France over five days, with each student having one of 9 labels representing classes. We divide one day into 10 intervals, so there are 50 snapshots in total. The initial snapshot has 279 nodes and 791 edges, and the final snapshot has 76 nodes and 80 edges. 
    \item \textbf{DBLP} dataset is a co-author network in the computer science field. The label of an author is defined by the fields in which the author has the most publications, and there are 15 labels in total. The 8 snapshots (1985-1992) are taken out to form this dynamic network. The initial snapshot has 1679 nodes and 3445 edges, and the final snapshot has 12107 nodes and 25841 edges. For \textbf{DBLP$_{dyn}$} dataset, we take 10 snapshots (2010-2019) and remove nodes with unknown labels. Note that DBLP is an accumulative network where nodes and edges do not disappear once they are added. In contrast, the coauthor communities of DBLP$_{dyn}$ appear and disappear dynamically.
    \item \textbf{Cora} dataset is a citation network where each node represents a paper, and an edge between two nodes represents a citation. Each paper is assigned a label (from 10 different labels) based on its field of publication. The 11 snapshots (1989-1999) are taken out to form its dynamic network. The initial snapshot has 348 nodes and 481 edges, and the final snapshot has 12022 nodes and 45421 edges. 
\end{itemize}

\begin{table}[htbp]
\resizebox{0.48\textwidth}{!}{
\begin{tabular}{lllllll}
\hline
Dataset    & $\left| V\right|_{min}$ & $\left|E\right|_{min}$ & $\left| V\right|_{max}$  & $\left|E\right|_{max}$  & $T$  & $K$  \\ \hline
Enron      & 18   & 22   & 166   & 677   & 15 & 7  \\
Highschool & 76   & 80   & 289   & 1214  & 50 & 9  \\
Cora       & 348  & 418  & 12022 & 45421 & 11 & 10 \\
DBLP       & 1679 & 3445 & 12107 & 25841 & 8  & 15 \\
DBLP$_{dyn}$  & 2450 & 6460 & 13738 & 41934 & 10 & 14 \\ \hline
\end{tabular}
}
\caption{Dataset summary}
\label{table.dataset}
\end{table}

\section{Additional Experimental Results}
\begin{figure}[htbp] 
\centering 
\includegraphics[width=0.48\textwidth]{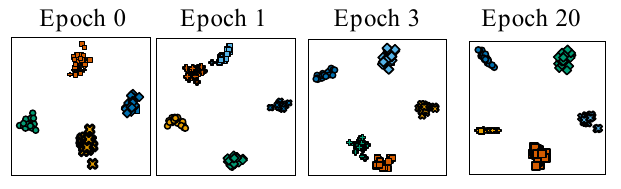} 
\caption{Evolution of graph embedding when Wasserstein distance of community topology between two demo graphs decreases. The colors represent the clustering results and the marker represents the ground truth labels.} 
\label{Fig.emb_loss_decrease} 
\end{figure}

\begin{figure}[]
\centering 
\includegraphics[width=0.45\textwidth]{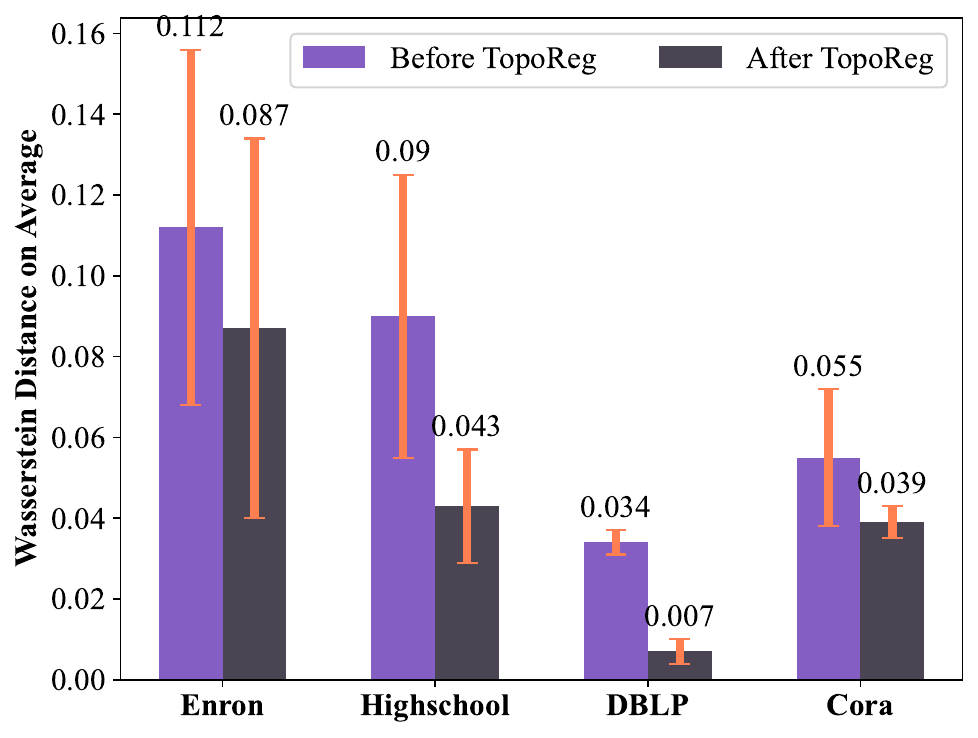} 
\caption{Wasserstein distance between community topologies in neighboring snapshots. The results prove that inter-community structure stability is improved on all four datasets after optimization using TopoReg.} 
\label{Fig.topoimprove} 
\end{figure}

\noindent\textbf{Fig. \ref{Fig.emb_loss_decrease}} shows the evolution of node embedding during the training process of TopoReg. The colors represent the clustering results and the marker represents the ground truth labels. The squares and vertical crosses are initially erroneously clustered in the orange cluster, and as the epoch increases, the wasserstein distance loss decreases, and the nodes represented by the squares and crosses gradually spread out and split into two communities.

\begin{figure*}[]
\centering 
\includegraphics[width=0.95\textwidth]{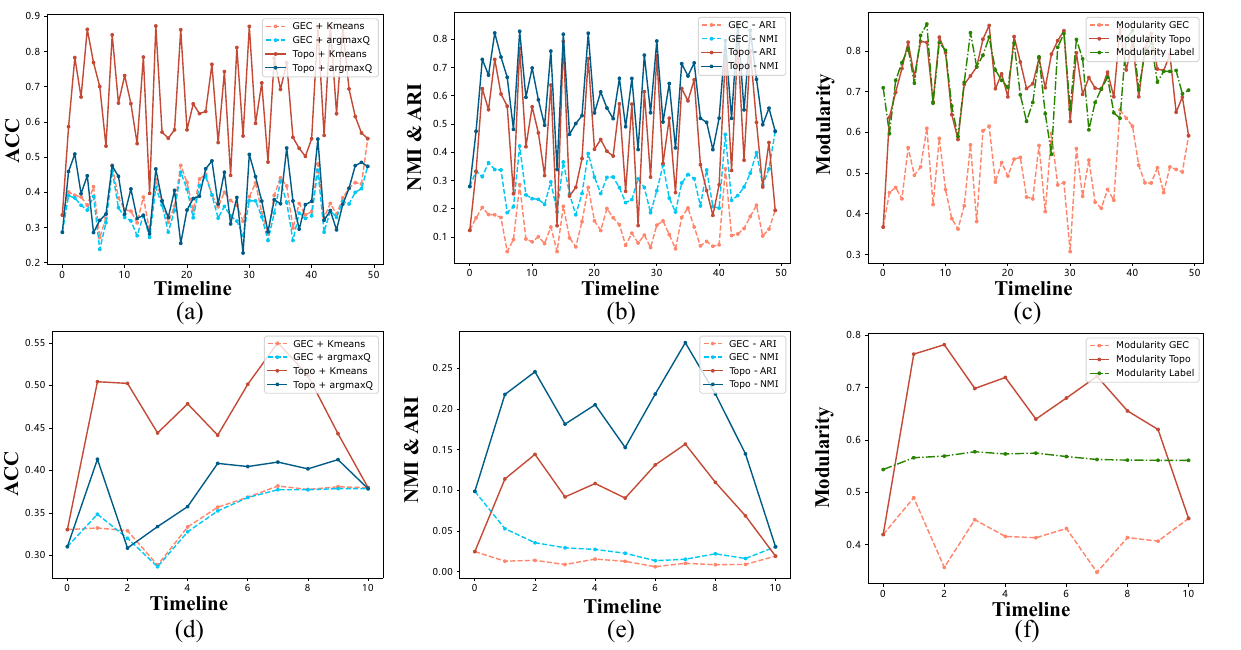} 
\caption{Metrics at each snapshots. The first row shows results on Highschool dataset while the second row is on Cora dataset.} 
\label{Fig.metrics} 
\end{figure*}

\noindent\textbf{Fig. \ref{Fig.metrics}} shows different evaluation metrics on each snapshots. The first row shows results on Highschool dataset while the second row is on Cora dataset. 

\noindent\textbf{Fig. \ref{Fig.topoimprove}} shows the mean and standard deviation of wasserstein distance between community topologies in neighboring snapshots. It can be seen that the mean and the standard deviation have varying degrees of decreases on all four datasets. The results prove that inter-community structure stability is improved after optimization applying TopoReg.

\section{Preliminary Knowledge of Topological Data Analysis}

Topological Data Analysis (TDA) is a rising research area that uses algebraic topology concepts to analyze complex, high dimensional data. Persistent homology is the fundamental methodology in TDA that summarizes the lifetimes of topological features within a filtration as a persistence diagram. We briefly introduce it and refer the readers to \cite{dey2022computational} for more details.

\subsection{Homology}
The key concept of homology theory is to study the properties of an object \(X\), such as a \(p\)-simplex, by means of commutative algebra. A geometric \(p\)-simplex is a convex combination of \(p+1\) (affinely) independent points in \(R^N\). Complex \(K\) is a collection of simplices. A \(p\)-chain is a linear combination of \(p\)-simplices under \(\mathbb{Z}_2\)-coefficients. For a simplex \(\sigma = [x_0, \dots, x_p] \in K\), we define boundary operators as \(\partial_p(\sigma) = \sum_{i=0}^{p}[x_0, \dots , x_{i - 1}, x_{i+1}, \dots , x_p]\) and linearly extend this to chain group \(C_p(K)\), i.e. \(\partial_p( \sum \sigma_i) = \sum \partial_p(\sigma_i)\). In particular, we assign to \(K\) a sequence of chain groups \(C_0, C_1, \dots\) , which are connected by boundary operators \(\partial_n : C_n  \rightarrow C_{n-1}\) such that \(\operatorname{im} (\sigma_{n+1}) \subset \operatorname{ker} (\sigma_n)\). By studying its homology groups \(H_n = \operatorname{ker} (\sigma_n)/\operatorname{im} (\sigma_{n+1})\), we can derive topology properties of \(K\). In this case, the ranks of homology groups yield directly interpretable properties, e.g. \(rank(H_0)\) reflects the number of connected components and \(rank(H_1)\) the number of loops. We call it betti number \(\beta_p(K)=dim(H_p) = rank(Z_p)-rank(B_p)\).


\subsection{Persistent Homology}
Let $\left(K^i\right)_{i=0}^m$ be a sequence of simplicial complexes such that $\emptyset=K^0 \subseteq K^1 \subseteq \cdots \subseteq K^m=K$. Then, $\left(K^i\right)_{i=0}^m$ is called a filtration of $K$. Using the extra information provided by the filtration of $K$, we obtain a sequence of chain complexes where $C_p^i=C_p\left(K_p^i\right)$. This leads to the concept of persistent homology groups, i.e.,

$$
H_p^{i, j}=\operatorname{ker} \partial_p^i /\left(\operatorname{im} \partial_{p+1}^j \cap \operatorname{ker} \partial_p^i\right) \quad \text { for } \quad 1 \leq i \leq j \leq m .
$$

The ranks, $\beta_p^{i, j} = rank(H_p^{i, j})$, of these homology groups (i.e. the $p$-th persistent Betti numbers), capture the number of homological features of dimension $p$ that persist from $i$ to (at least) $j$. we can construct a multiset by inserting the point $(a_i , a_j ), 1 \leq i < j \leq m$, with multiplicity $\mu_p^{i, j}=\left(\beta_p^{i, j-1}-\beta_p^{i, j}\right)-\left(\beta_p^{i-1, j-1}-\beta_p^{i-1, j}\right)$ . This effectively encodes the $p$-dimensional persistent homology of $K$ w.r.t. the given filtration. This representation is called a persistence barcode $B_p$, or in the form of a diagram called persistence diagram $D_p$. There are many metrics for  persistent homology, and one of them, called Wasserstein distance, is defined as the optimal transport distance between the points of the two diagrams:
\subsection{Wasserstein distance}
    For $p>0$, the $p$-Wasserstein distance between two persistence diagrams $D_k^{(1)}$ and $D_k^{(2)}$ is defined
$$
\mathrm{W}_{p, q}\left(D_k^{(1)}, D_k^{(2)}\right)=\inf _{\gamma: D_k^{(1)} \rightarrow D_k^{(2)}}\left(\sum_{x \in D_k^{(1)}}\|x-\gamma(x)\|_q^p\right)^{1 / p},
$$
    where $\|\cdot\|_q$ denotes the $q$-norm, $1 \leq q \leq \infty$ and $\gamma$ ranges over all bijections between $D_k^{(1)}$ and $D_k^{(2)}$. 
\subsection{Weight Rank Clique Filtration}
Weight rank clique filtration (WRCF) \cite{petri2013topological} prioritizes the addition of edges with larger weights before those with smaller weights to emphasize the importance of edge weights. Specifically, the WRCF algorithm progressively creates unweighted graphs at each filtration step and extracts maximal cliques as simplices for further study. After obtaining a filtered complex on which we can subsequently apply persistent homology. This approach presents one of the feasible methods for applying persistent homology to temporal networks \cite{lozeve2018topological}.

\subsection{Gradient of Persistence Homology}
Substantial research advancements have been made in the field of topological optimization targeting persistence homology \cite{brüelgabrielsson2020topology,vandaele2022topologically}. Given an input filtration $f: \mathcal{X} \rightarrow$ $\mathbb{R}$, we can compute the gradient of a functional of a persistence diagram $\mathcal{E}\left(\mathrm{dgm}_k\right) =\mathcal{E}\left(\{ (b_i, d_i) \}_{i \in I_k}\right)
$ by mapping each birth-death pair to the cells that respectively created and destroyed the homology class, defining an inverse map
\begin{equation}
\label{eq:inv map}
\pi^k_f:\left\{b_i, d_i\right\}_{i \in I_k} \rightarrow(\sigma, \tau) .
\end{equation}
In the case where the ordering on simplices is strict, the map is unique, and we compute the gradient as:
\begin{equation}
\frac{\partial \mathcal{E}}{\partial \sigma}=\sum_{i \in I_k} \frac{\partial \mathcal{E}}{\partial b_i} \mathbf{1}(\pi^k_f\left(b_i\right)=\sigma)+\sum_{i \in I_k} \frac{\partial \mathcal{E}}{\partial d_i} \mathbf{1}(\pi^k_f\left(d_i\right)=\sigma)
\end{equation}
in which at most one term have a non-zero indicator. With the inverse map, we can skip the derivation of the topological calculations and directly optimize the filtration value of the corresponding simplex, which is the output of clustering module in the neural network in our method. Based on this concept, this paper extends the optimization of persistence homology to the scenario of graph filtration.



\bibliography{supplement}